\documentclass[sigconf,authorversion,nonacm]{acmart}

\usepackage{multicol}
\usepackage{multirow}
\usepackage{colortbl}
\usepackage{subcaption}
\usepackage{hyperref}
\usepackage{lipsum}
\newcommand\blfootnote[1]{%
  \begingroup
  \renewcommand\thefootnote{}\footnote{#1}%
  \addtocounter{footnote}{-1}%
  \endgroup
}

\usepackage{dblfloatfix}

\AtBeginDocument{%
  \providecommand\BibTeX{{%
    \normalfont B\kern-0.5em{\scshape i\kern-0.25em b}\kern-0.8em\TeX}}}

\begin{document}

\title[The mapKurator System (Demo Paper)]{The mapKurator System: A Complete Pipeline for Extracting and Linking Text from Historical Maps (Demo Paper)}

\author{Jina Kim, Zekun Li, Yijun Lin, Min Namgung, Leeje Jang, Yao-Yi Chiang}
\email{[kim01479, li002666, lin00786, namgu007, jang0124, yaoyi]@umn.edu}
\affiliation{%
  \institution{University of Minnesota}
  \city{Minneapolis}
  \country{USA}
}







\renewcommand{\shortauthors}{Kim et al.}

\begin{abstract}

Scanned historical maps in libraries and archives are valuable repositories of geographic data that often do not exist elsewhere. Despite the potential of machine learning tools like the Google Vision APIs for automatically transcribing text from these maps into machine-readable formats, they do not work well with large-sized images (e.g., high-resolution scanned documents), cannot infer the relation between the recognized text and other datasets, and are challenging to integrate with post-processing tools. This paper introduces the \textit{mapKurator} system, an end-to-end system integrating machine learning models with a comprehensive data processing pipeline. \textit{mapKurator} empowers automated extraction, post-processing, and linkage of text labels from large numbers of large-dimension historical map scans. The output data, comprising bounding polygons and recognized text, is in the standard GeoJSON format, making it easily modifiable within Geographic Information Systems (GIS). The proposed system allows users to quickly generate valuable data from large numbers of historical maps for in-depth analysis of the map content and, in turn, encourages map findability, accessibility, interoperability, and reusability (FAIR principles). We deployed the \textit{mapKurator} system and enabled the processing of over 60,000 maps and over 100 million text/place names in the David Rumsey Historical Map collection. We also demonstrated a seamless integration of \textit{mapKurator} with a collaborative web platform to enable accessing automated approaches for extracting and linking text labels from historical map scans and collective work to improve the results.


\end{abstract}


\maketitle

\section{Introduction}

\begin{figure*}[!b]
  \centering
  \includegraphics[width=0.9\linewidth]{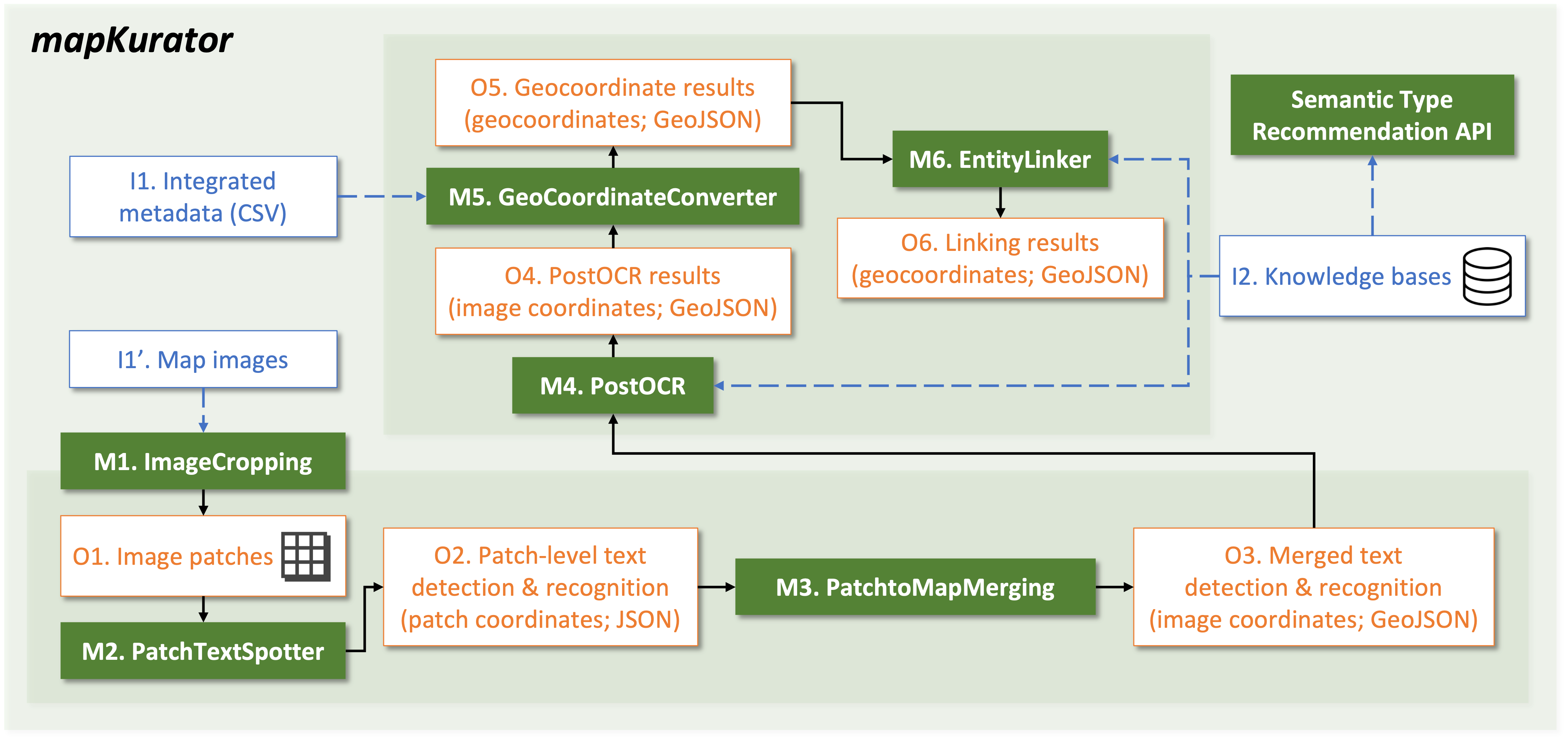}
  \caption{System workflow with \textit{mapKurator} architecture; \textit{mapKurator} consists of six modules (abbreviated as M\#) and semantic type recommendation API. O\# denotes output files and I\# with dotted lines are the inputs for each module.}\label{fig:workflow}
\end{figure*}

\blfootnote{\textit{This is the author's version of the work and it is posted for personal use.}}
Historical maps constitute an extensive body of geographic information providing a detailed understanding of changes over time~\cite{rumsey2002historical}. Many historical maps are publicly available from online map archives and digital libraries~\cite{chiang2020using}. One of the largest collections of scanned historical maps is the David Rumsey Map Collection, which has more than 115,000 scanned maps from the 16th through the 21st century.~\footnote{\url{https://www.davidrumsey.com/}} To enable advanced search queries allowing users to retrieve relevant historical maps, data curators of various backgrounds, including geographers, historians, and librarians, have put substantial efforts into generating comprehensive metadata for individual map scans using map content and additional data sources. However, creating and maintaining metadata of scanned maps requires expert knowledge and extensive manual work~\cite{rumsey2002historical}. 

There have been many attempts to convert text labels on maps to analytic-ready data, including crowdsourced projects (e.g., GB1900~\footnote{\url{https://geo.nls.uk/maps/gb1900/}}) and machine learning model-based approaches~\cite{li2020automatic, weinman2019deep, li2018intelligent}. Although prior studies have actively investigated automatic approaches, they focus on developing machine learning models for text detection and recognition with limited emphasis on the end-to-end data pipeline from large numbers of map images to repositories of searchable text labels. Also, the automated results can still be imperfect, leading to inaccurate map metadata, discouraging the user community, and creating a gap between end-users and technologies. For example, a user using the Google Vision API or a text spotting model will need to take care of large input image dimensions, write specialized code for handling geocoordinates and format conversion, and often require to conduct ad-hoc post-processing to improve the automated results. These steps require particular technology skills (e.g., writing Python scripts for data join and format conversion) and IT (information technology) resources, which might not be available for a wide variety of users and institutes. This human-computer gap limits the broader impact of the automatic map processing approaches and hinders historical maps' findability, accessibility, interoperability, and reusability (FAIR principles~\cite{wilkinson2016fair}). 

To bridge the gap between end-users and intelligent technologies and promote the FAIR principles of historical maps, we present the \textit{mapKurator} system that allows users to leverage ready-to-use automatic map processing technologies for processing large numbers of large-dimension historical map scans to convert their text content into a standard, machine-readable format. \textit{mapKurator}'s overall system capabilities include automatic processes of 1) detecting and recognizing text from map scans of large dimensions, 2) automatic post-processing and linking text labels to their corresponding entities in external knowledge bases, and 3) recommending semantic types of user's input. Target users of the proposed system include historians, geographers, librarians, and other researchers who study full-textual content on historical maps or use the text label to generate map metadata (e.g., geographical things described in the text label, their locations, and their semantics and other attributes).

We have built and publicly released Docker images of the \textit{mapKurator} system\footnote{\url{https://knowledge-computing.github.io/mapkurator-doc/}} and the integration of the \textit{mapKurator} system with a 
collaborative web-based annotation platform (Recogito).\footnote{\url{https://recogito.pelagios.org/}} The deployment of the \textit{mapKurator} system at the University of Minnesota has processed over 60,000 maps and over 100 million text/place names in the David Rumsey Historical Map collection. These 100 million text labels have been incorporated into the metadata platform to support a full-text search of map content by Luna Imaging.\footnote{\url{https://mailchi.mp/stanford/apr2023-ai-advancements-in-map-studies}} The \textit{mapKurator} system has also processed thousands of Sanborn Fire Insurance maps from the Library of Congress and historical Ordinance Survey maps from the National Library of Scotland. We are in the process of releasing the processing results as open research data.

The following sections describe the workflow of the \textit{mapKurator} system and its architecture, related work on annotation interfaces and text spotters, and a discussion on future work.

\section{The \textit{mapKurator} System}

Figure~\ref{fig:workflow} shows the proposed \textit{mapKurator} system. The data pipeline is as follows. \textit{mapKurator} takes a map image as input, slices the image into small patches (M1), detects and recognizes text labels on each patch (M2), combines the predicted text labels from all patches (M3), performs automatic post-processing (M4)\cite{min22}, converts their image coordinates to geocoordinates (M5), and links to the entities in external knowledge bases (M6). The final output is in the GeoJSON format, which can be viewed and edited across various geographic information systems (e.g., QGIS). The user can also collaboratively modify \textit{mapKurator} results on the \textit{Recogito} web interface using the integrated \textit{mapKurator} and \textit{Recogito}. The details of each module and GeoJSON format are as follows.


\textbf{M1. ImageCropping} High-resolution historical map images typically contain a large amount of detail, which can result in very large file sizes. This poses a challenge when working with the images due to the large memory requirement. To overcome this challenge, the system crops the map images into smaller tiles, where each tile contains a sub-region (e.g., 1000 $\times$ 1000 pixels) of the input. 

\textbf{M2. PatchTextSpotter} \textit{mapKurator} utilizes a text-spotting model to detect and recognize text on the image patches. Our model builds upon Deformable DETR~\cite{zhu2020deformable} and TESTR~\cite{zhang2022text}, in which the encoder processes the image features and generates text proposals (i.e., coarse bounding boxes), while the decoder refines the proposals to obtain arbitrary-shaped detection results and extracts text within the proposals. To train the text spotter effectively, we employ synthetic datasets, including a synthetic scene image dataset~\cite{gupta2016synthetic}, a novel synthetic map dataset designed to mimic text placement on historical maps, inspired by~\cite{li2021synthetic}, along with real-world images with human annotations. The output of this module includes the detected text regions, comprising 16 boundary points with image patch coordinates and the corresponding recognized text.

\textbf{M3. PatchtoMapMerging} After performing text spotting on each tile independently in parallel, \textit{mapKurator} merges the results from each tile by collecting all the patch-level predictions and shifting the predicted image coordinates according to the patch location. 

\textbf{M4. PostOCR} To improve the output of text spotting results, \textit{mapKurator} runs lexical-based post-OCR processing by using the edit distance to compare the text spotting results and a vocabulary set. \textit{mapKurator} generates the vocabulary set from geo-entities in OpenStreetMap and breaks ties using the geo-entity's popularity (i.e., frequency).

\textbf{M5. GeocoordinateConverter} \textit{mapKurator} takes the map metadata, which includes ground control point pairs and transformation methods (e.g., the affine transformation) to convert the predicted bounding polygons (i.e., spotting results) from image coordinates to geocoordinates by using translator library, GDAL\footnote{\url{https://gdal.org/index.html}}. 

\textbf{M6. EntityLinker} After the coordinate transformation, \textit{mapKurator} links the post-processed text label to the corresponding geo-entities in external knowledge bases (e.g., OpenStreetMap, Wikidata) or historical gazetteers. The identified linkages enable advanced search queries on scanned maps by leveraging geo-entity properties in knowledge bases~\cite{li2020automatic}. In the current version of the \textit{mapKurator} system, EntityLinker retrieves the candidate geo-entities in OpenStreetMap that satisfy two criteria: 1) the suggested word (i.e., output from M4) is a substring of the candidate geo-entity's name and 2) the geocoordinates of text bounding polygon (i.e., output from M5) is in the geometry of OpenStreetMap geo-entities.

\vspace{-.01in}
\begin{figure}[h]
    \vspace{-.12in}
    \centering
    \includegraphics[width=0.92\linewidth]{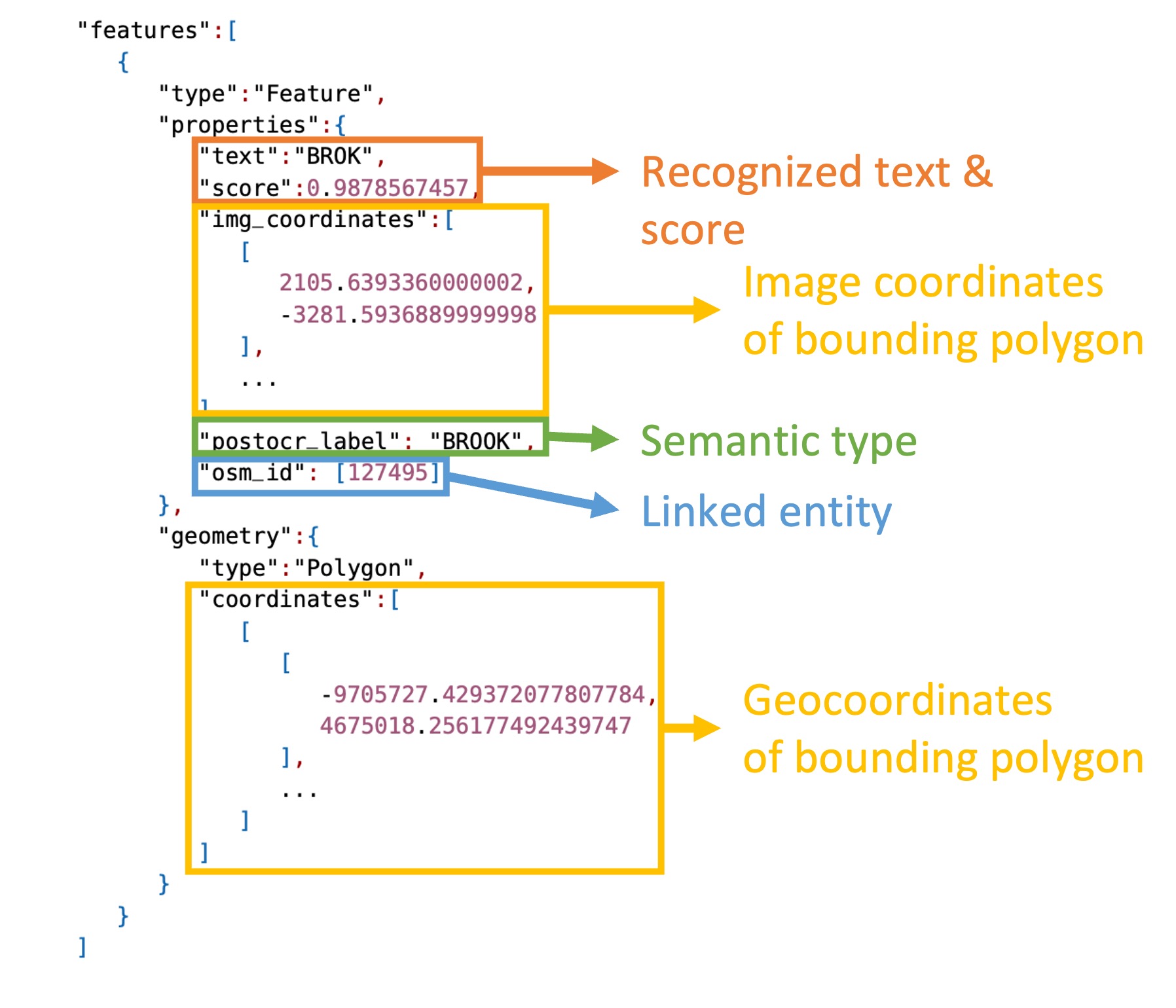}
    \vspace{-.15in}
    \caption{A text label example of the \textit{mapKurator}'s output file in GeoJSON format.}\label{fig:output}
    \vspace{-.1in}
\end{figure}

Figure~\ref{fig:output} shows an example GeoJSON result that refers to one of the processed text via \textit{mapKurator}. Each GeoJSON feature contains the geocoordinates of the bounding polygon (the `coordinates' field in Figure~\ref{fig:output}) and several properties, including the predicted text label with score (`text' \& `score' field), post-processed text label (`postocr\_label' field), and the unique identifiers of matched entities in the external knowledge bases. The post-processed text `BROOK' has the linked entity from OpenStreetMap (`osm\_id' field). The system can also match the text label in other external knowledge bases or historical gazetteers.

\begin{figure}[h]
\vspace{-.14in}
  \centering
  \includegraphics[width=0.7\linewidth]{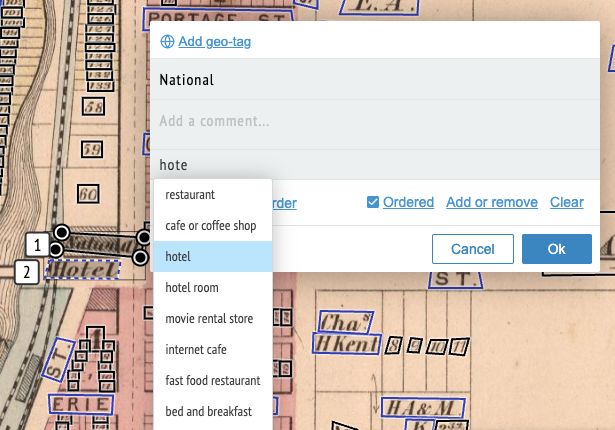}
  \caption{A list of semantic types retrieved from recommendation API while typing `hote'.}\label{fig:semantic_type}
  \vspace{-.15in}
\end{figure}

To demonstrate \textit{mapKurator}'s results and facilitate interoperability and findability of the text on maps, the \textit{mapKurator} system also provides a semantic type recommendation application programming interface (API),\footnote{\url{https://github.com/machines-reading-maps/semantic-type-recommendation-api}} for semi-automatic type selection in the integrated \textit{mapKurator} and Recogito. Traditional approaches for finding the relevant semantic type for a text on the map from a large set of vocabulary are based on expert efforts and are time-consuming. The API employs a fastText model pretrained on large corpora of Wikipedia~\cite{bojanowski2016enriching} with 240 standardized semantic types from Schema.org~\footnote{\url{https://schema.org}} to retrieve semantically semantic types in real-time while typing (Figure~\ref{fig:semantic_type}). 

\begin{figure*}[h]
    \centering
    \includegraphics[width=\linewidth]{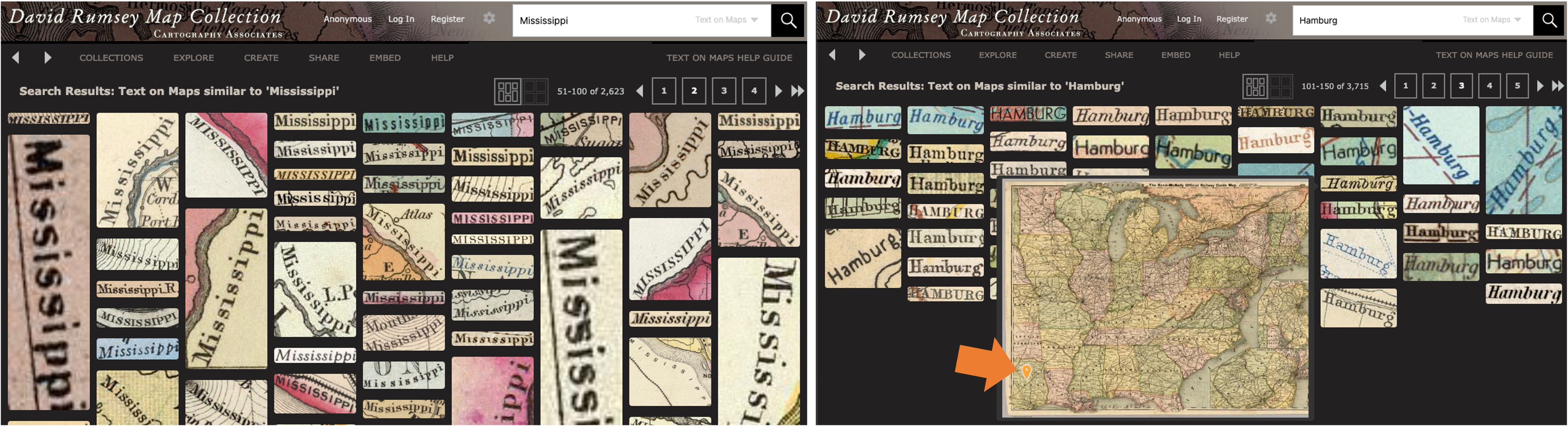}
    \caption{Visualizations of \textit{mapKurator}’s results from the David Rumsey Map Collection (web interface powered by Luna Imaging). The left figure presents the search results for `Mississippi', and the right figure shows the search results for `Hamburg'. Each tile represents a detected and recognized text label from the 60,000 processed georeferenced maps. Note that these text labels can be in varying orientations and font sizes. A toggled text label on the right figure shows the full map image with pinned location (i.e., orange arrow).}\label{fig:luna}
\end{figure*}

In summary, the \textit{mapKurator} system provides ready-to-use intelligent technologies, allowing users to extract and link text on maps. The system also promotes the four fundamental FAIR principles for historical maps. The results enable advanced search queries for finding relevant historical maps. Figure~\ref{fig:luna} shows examples of \textit{mapKurator}'s results from Luna Imaging.

\vspace{-.1in}
\section{Related Work}


Extracting text from scanned images rely on manual annotations or text-spotting tools. One of the most common image labeling software tools called LabelMe~\cite{russell2008labelme} has discussed the potential semi-automatic approaches to leveraging image processing algorithms and online image search engines to assist manual labeling of images. The user of LabelMe can validate a predicted bounding box and edit the label by removing or redrawing the bounding box. Another interactive annotation tool, iVAT~\cite{bianco2015interactive}, is designed for video annotation and supports three different annotation approaches, manual, semi-automatic, and automatic. Regarding semi-automatic annotation, iVAT requires users to annotate an initial set of given video frames and applies automatic algorithms to provide the remaining annotations. However, the aforementioned systems mainly focus on generating annotated data for training machine learning models and do not consider these annotated data as spatial things (e.g., see cartographic interaction~\cite{roth2012cartographic}), such as text on maps. In contrast, Recogito allows users to collaboratively annotate text bounding boxes in diverse shapes and transcribe the text but still, the process is time-consuming and does not scale to process large numbers of maps. State-of-the-art text spotting models, such as TESTR~\cite{zhang2022text} and SwinTextSpotter~\cite{huang2022swintextspotter}, detects a bounding polygon and recognizes the text for each text instance in end-to-end trainable approach. Existing text spotters typically focus on scene images such as advertisements and rescale an input image into a small, fixed-size image, which cannot be directly applied to historical maps. Improving text spotters to address complex historical maps with a variety of cartographic styles requires lots of programming skills and additional steps to exploit map geocoordinates for refining the results. 

To process a large number of historical maps and provide an easily transferable standardized output, \textit{mapKurator} system proposes a complete end-to-end data pipeline with well-defined modules. The system makes it easy for users to utilize the latest development in text spotter for extracting text on maps. Also, the integration of the \textit{mapKurator} system and Recogito enables ready-to-use intelligent technologies with a collaborative annotation web interface, allowing the generation of complete text content or metadata of historical maps for users of all levels of technical skills. 

\vspace{-.15in}
\section{Discussion and Future Work}
We presented a complete map processing system, \textit{mapKurator}, that consists of ready-to-use intelligent technologies and the integration of the map processing system with a collaborative web-based annotation platform. \textit{mapKurator} fills the gap between users and automatic approaches and promotes the fundamental four FAIR principles for historical maps and their content. 
We plan to incorporate the capability in \textit{mapKurator} to process maps with multiple languages.
We also plan to build interactive machine-learning approaches, exploiting users' annotations to actively improve the results. Moreover, we will support various ontologies and file formats to address the demands of wide user communities. 

\vspace{-.1in}
\begin{acks}
This material is based upon work supported in part by NVIDIA Corporation, the gift from David and Abby Rumsey to the University of Minnesota Foundation, and the University of Minnesota, Computer Science \& Engineering Faculty startup funds. We thank Tanisha Shrotriya and Rainer Simon for their efforts in the integration of the \textit{mapKurator} system and Recogito. We especially thank David Rumsey for his generous support and encouragement. 
\end{acks}


\vspace{-.1in}
\bibliographystyle{ACM-Reference-Format}
\bibliography{main}


\end{document}